# 全向驱动旋翼机械臂设计、建模及 dc-PID 控制

马　乐，王　东，郝子绪，孙加森，刘　杰，王诗语

（东北电力大学 自动化工程学院 机器人技术实验室，吉林 吉林 132012）

**摘　要**：提出了一种新型全向驱动旋翼机械臂的设计、建模与基于动态补偿 PID 旋翼平台控制方法。首先设计并分析了新型全向驱动旋翼机械臂机构；然后采用 Craig 参数法与递归 Newton-Euler 方程建立系统运动学与动力学模型，得出了旋翼平台与机械臂间精确动力学关系；针对旋翼平台全向驱动控制问题提出了动态补偿 PID 控制方法，最后推导出统一的多旋翼平台耦合矩阵方程实现设计与解耦计算。仿真考虑了随机噪声与扰动、参数不确定性、功率限制等实际问题，结果得出本文设计的系统能完成传统旋翼机械臂理论上无法完成的全向驱动控制并具有良好的控制品质，与 **Backstepping** 方法比较得出，在具有机械臂运动的任务级控制中，动态补偿 **PID** 控制有更好的控制精度和更强的扰动抑制能力。动态补偿控制实验验证了本文方法的实用性与有效性。

**关键词**：旋翼机械臂；全向驱动；动力学建模；补偿 PID 控制；倾角悬停

**中图分类号**：TP242.3　　　　　　　　　　**文献标识码**：A

# Design，Modeling and Dynamic Compensation PID Control of a Fully-Actuated Aerial Manipulation System

MA Le, WANG Dong, HAO Zixu, LIU Jie, SUN Jiasen, WANG Siyu

（*Robotics Technology Lab, School of Automation and Engineering, Northeast Electric Power University,,Jilin 132012, China*）

**Abstract:** This paper addresses design, modeling and dynamic-compensation PID (dc-PID) control of a novel type of fully-actuated aerial manipulation (AM) system. Firstly, design of novel mechanical structure of the AM is presented. Secondly, kinematics and dynamics of AM are modeled using Craig parameters and recursion Newton-Euler equations respectively, which give rise to a more accurate dynamic relationship between aerial platform and manipulator. Then, the dynamic-compensation PID control is proposed to solve the problem of fully-actuated control of AM. Finally, uniform coupled matrix equations between driving forces/moments and rotor speeds are derived, which can support design and analysis of parameters and decoupling theoretically. It is taken into account practical problems including noise and perturbation, parameter uncertainty, and power limitation in simulations, and results from simulations shows that the AM system presented can be fully-actued controlled with advanced control performances, which can not achieved theoretically in traditional AM. And with compared to backstepping control dc-PID has better control accuracy and capability to disturbance rejection in two simulations of aerial operation tasks with motion of joint. The experiment of dc-pid proves the availability and effectiveness of the method proposed.

## 1 引言（Introduction）

近年，多旋翼无人机因其灵活高效系统已得到较深入研究和广泛应用。然而传统多旋翼无人机不具有执行结构，仅能完成监测、航拍等被动任务。将机械臂与多旋翼结合的旋翼机械臂(Aerial Manipulation，AM)系统能够弥补执行能力的不足[1]，能应用于诸如空中操作、设施维护、抢险救援等广泛的主动任务领域。因此，旋翼机械臂系统是必然的发展方向[2]。

旋翼机械臂系统存在如下几方面复杂性[3]：1）结构属树形无根多刚体系统，具有强耦合性并对负载敏感；2）系统非完整并欠驱动；3）存在子系统间内力/矩动态约束与外力/矩扰动。其设计、建模与控制等问题是极具挑战性的关键课题。以至近年众多国内外学者针对其中问题展开研究。以下从机构设计、系统建模与控制方法三方面分析研究现状。

已提出了不同结构的旋翼机械臂系统，其中涵盖了不同飞行平台（包括直升机[4]、四旋翼[5]、六旋翼[6]、八旋翼[7]）与不同结构机械臂（包括单轴、多轴[8]、delta[9]与双臂[10]等）的组合方式。相比于 delta 结构，串联机械臂工作空间更大运动更灵活，然而串联机械臂的自由度选择存在灵活性、复杂性与负载能力三者的平衡问题[11]。文献[12]指出较直升机平台，多旋翼平台驱动耦合性更低且具更好的灵活性。目前的多旋翼平台多为非完整的，旋翼转轴平行或有微小角度[2]，仅能产生相对机体坐标系驱动力，以至无法实现全向驱动控制，即位置控制

必牺牲姿态控制精度(反之亦然)，限制了空中作业灵活性。文献[13]采用 7 自由度冗余机械臂以提升操作灵活性，但提高了机械臂与平台协同难度，降低了负载能力。

目前多采用 Euler 角[14]与 D-H[15]参数法描述旋翼平台与机械臂运动学。但 AM 系统动力学建模存在差异[1]。研究采用的数学工具虽集中于 Newton-Euler[19]与 Lagrange[16]动力学方程，但因建模复杂多以不同角度简化处理。文献[6, 17, 18]等将 AM 系统简化为单刚体，采用动态惯性张量补偿模型简化的精度。然而该方法在动量定理中的理论依据与精度有待讨论，且从算法过程看出计算动态惯性张量与质心位置本身难度巨大。文献[19]仅联立了机械臂二阶动力学与 New-Euler 法的旋翼平台动力学方程，未说明内力计算方法。[20]等则视 AM 系统交互内力(矩)为输入，但实际系统中该条件难以实现。此外[21]等将动力学降至 2 维建模，但适用范围有限。综上 AM 系统实用有效的动力学建模问题扔亟待解决。

众多控制方法均被采用于 AM 系统[22]，但主要集于 PID[13, 23]、Backstepping[24]等。除 PID 外大多控制方法均依赖精确模型。由于 AM 系统存在强非线性、动态约束及上述动力学建模问题，已有方法特别在噪声扰动下的控制效果有待提升。文献[25]采用增强学习方法提升控制性能，但学习控制方法时间成本较大且其实用仍需以稳定的控制方法为前提。综上，目前 AM 系统控制的核心应着眼于对强非线性与动态约束的处理问题。

综上三方面论述，结合作者前期工作[3]，并受[26]启发，本文设计了一种新型全向驱动旋翼机械臂系统结构，建立了动力学模型，提出了动态补偿 PID 控制方法以实现全向驱动控制。本文主要工作与创新贡献如下：

1）设计并建立了一种新型大倾角旋翼平台，以实现旋翼平台的全向驱动功能；

2）基于递归 Newtow-Euler 方程，建立了更适用于 AM 系统的内力（矩）动力学模型，使能更精确估计机械臂与旋翼平台作用力（矩）；

3）推导出统一的旋翼平台驱动力耦合矩阵公式，有利于参数分析与优化；

4）加入动态内力（矩）补偿项于 PID 控制，实现全向驱动控制，使旋翼平台达到更好性能同时实现了传统 AM 系统理论上无法完成的带有机械臂运动的倾角悬停控制。

## 2 全向 AM 系统机构设计（Mechanical Design of Fully-Actuated AM System）

图 1 为本文设计的全向驱动 AM 系统的 SolidWorks 结构图。不同于传统 AM 系统的旋翼平台本文采用大倾角旋翼设计。如图 2 所示，旋翼装配位置相对旋翼平台坐标系 x、y 轴对称，且旋翼与 x-y 平面夹角 $\beta_i$ 相等且<60°。设计较大旋翼倾角能产生相对旋翼平台 x 与 y 方向的力(矩)，进而理论上具有全向驱动能力，**该能力是传统 AM 系统不具备的**。

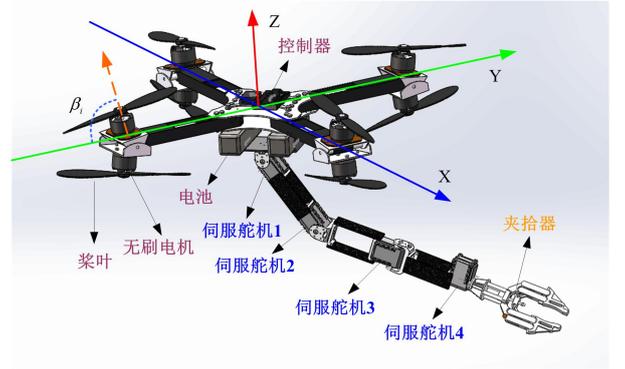

图 1 全向驱动 AM 系统 SolidWorks 结构图
Fig.1 Structure chart fully-actuated AM system in SolidWorks

考虑操作灵活性兼顾负载能力，本文采用 4 自由度机械臂结构。相比低自由度机械臂，配合全向旋翼平台的 4 自由度机械臂具有较高的操作灵活性，相较于高自由度机械臂 4 自由度又具有更高的负载能力。本文选用 Dynamixel AX-12A 伺服舵机作为关节驱动器[27]，该款电机具有位置、速度等反馈功能，并能以位置、速度作为指令控制关节运动。本文采用碳纤维结合 3D 打印技术设计加工样机，以降低机体自重同时便于优化设计。

## 3 系统建模（System Modeling）

### 3.1 运动学建模

设惯性参考坐标系、旋翼平台坐标系和机械臂第 $i$ 关节坐标系分别为 $\{W\}$、$\{A\}$、$\{M_i\}$，$\{A\}$ 相对 $\{W\}$ 位姿表示为：

$$^0\mathbf{T}_A = \begin{pmatrix} ^0\mathbf{R}_A & ^0\mathbf{P}_A \\ \mathbf{0}_{1\times 3} & 1 \end{pmatrix} \quad (1)$$

其中 $^0\mathbf{P}_A$ 为 $\{A\}$ 原点在 $\{W\}$ 下的向量，$^0\mathbf{R}_A$ 为 $\{A\}$ 相对 $\{W\}$ 的旋转矩阵，

$$^0\mathbf{R}_A = R_x(\varphi)R_y(\psi)R_z(\gamma) \quad (2)$$

$\mathbf{e}_{euler} = [\varphi, \psi, \gamma]$ 为旋翼平台 x-y-z Euler 角。

机械臂关节坐标系 $i-1$ 与 $i$ 间变换矩阵表示为（当 $i=1$ 时，表示关节坐标系 1 与 $\{A\}$ 变换矩阵，且下文规定旋翼平台为连杆 0）：

$$^{i-1}\mathbf{A}_i = \begin{pmatrix} ^{i-1}\mathbf{R}_i & ^{i-1}\mathbf{P}_i \\ \mathbf{0}_{1\times 3} & 1 \end{pmatrix} \quad (3)$$

其中 $^{i-1}\mathbf{R}_i$ 相邻坐标系旋转矩阵，$^{i-1}\mathbf{P}_i$ 为 $i$ 系原点在 $i-1$ 下坐标。采用 Craig 法定义参数 $\langle \boldsymbol{\alpha}, \mathbf{a}, \mathbf{d}, \boldsymbol{\theta} \rangle$ 并计算 $^{i-1}\mathbf{A}_i$，

$$^{i-1}\mathbf{A}_i = \begin{pmatrix} c\theta_i & -s\theta_i c\alpha_i & s\theta_i s\alpha_i & a_i c\theta_i \\ s\theta_i & c\theta_i c\alpha_i & -c\theta_i s\alpha_i & a_i s\theta_i \\ 0 & s\alpha_i & c\alpha_i & d_i \\ 0 & 0 & 0 & 1 \end{pmatrix} \quad (4)$$

其中 $\langle\boldsymbol{\alpha},\mathbf{a},\mathbf{d}\rangle$ 为常数，$\boldsymbol{\theta}$ 为关节角度变量，由(4)看出 $^{i-1}\mathbf{P}_i$ 为常量。表 1 为本文设计的机械臂 Craig 配置参数表（为在 Craig 定义下方便表示末端执行器，加设关节 5）。

表 1 机械臂 Craig 参数表
Tab.1 Craig parameter ist of manipulator

| $\boldsymbol{\alpha}$ | $\mathbf{a}$ | $\mathbf{d}$ | $\boldsymbol{\theta}$ |
| --- | --- | --- | --- |
| 0 | 0 | 0 | $\theta_1$ |
| $l_1$ | 0 | 0 | $\theta_2$ |
| $l_2$ | $-\pi/2$ | 0 | $\theta_3$ |
| 0 | $-\pi/2$ | $l_3$ | $\theta_4$ |
| 0 | 0 | $l_4$ | $\theta_5 \equiv 0$ |

由坐标系间传递关系，可得各关节坐标系相对 $\{W\}$ 系位姿矩阵：

$$^{0}\mathbf{T}_i = \begin{pmatrix} ^{0}\mathbf{R}_i & ^{0}\mathbf{P}_i \\ 0 & 1 \end{pmatrix} = {^{0}\mathbf{T}_A}\prod_{k}^{i}{^{k-1}\mathbf{T}_k} \quad (5)$$

进而得到各坐标系原点与连杆质心在 $\{W\}$ 系表示，

$$\begin{cases} ^{0}\mathbf{P}_i = {^{0}\mathbf{P}_{i-1}} + {^{0}\mathbf{R}_{i-1}}\,{^{i-1}\mathbf{P}_i} \\ ^{0}\mathbf{P}_i^{c} = {^{0}\mathbf{P}_i} + {^{0}\mathbf{R}_i}\,{^{i}\mathbf{P}_i^{c}} \end{cases} \quad (6)$$

式中 $^{0}\mathbf{P}_i^{c}$ 为连杆 $i$ 质心在 $\{W\}$ 向量。

旋翼平台线速度 $^{0}\dot{\mathbf{P}}_A$ 与角速度 $^{0}\boldsymbol{\omega}_A$ 由旋翼驱动力改变，且 $^{0}\boldsymbol{\omega}_A$ 与欧拉角导数存在关系[28]：

$$\begin{bmatrix} \dot{\varphi} \\ \dot{\psi} \\ \dot{\gamma} \end{bmatrix} = \begin{bmatrix} 1 & \sin\varphi\tan\psi & -\cos\varphi\tan\psi \\ 0 & \cos\varphi & \sin\varphi \\ 0 & \dfrac{-\sin\varphi}{\cos\psi} & \dfrac{\cos\varphi}{\cos\psi} \end{bmatrix}{^{0}\boldsymbol{\omega}_A} \quad (7)$$

关节坐标系角速度来自平台转动与关节角速度，其递推关系为：

$$^{0}\boldsymbol{\omega}_i = {^{0}\boldsymbol{\omega}_{i-1}} + {^{0}\mathbf{R}_i}\dot{\boldsymbol{\theta}}_i \mathbf{Z}_0 \quad (8)$$

$\mathbf{Z}_0 = [0,0,1]^{\mathrm{T}}$。(6)式求导得坐标系间线速度关系，

$$\begin{cases} ^{0}\dot{\mathbf{P}}_i = {^{0}\dot{\mathbf{P}}_{i-1}} + {^{0}\dot{\mathbf{R}}_{i-1}}\,{^{i-1}\mathbf{P}_i} \\ \quad = {^{0}\dot{\mathbf{P}}_{i-1}} + {^{0}\boldsymbol{\omega}_{i-1}} \times {^{0}\mathbf{R}_{i-1}}\,{^{i-1}\mathbf{P}_i} \\ ^{0}\dot{\mathbf{P}}_i^{c} = {^{0}\dot{\mathbf{P}}_i} + {^{0}\dot{\mathbf{R}}_i}\,{^{i}\mathbf{P}_i^{c}} \\ \quad = {^{0}\dot{\mathbf{P}}_i} + {^{0}\boldsymbol{\omega}_i} \times {^{0}\mathbf{R}_i}\,{^{i}\mathbf{P}_i^{c}} \end{cases} \quad (9)$$

对(8)、(9)式求导得到坐标系间加速度关系，

$$\begin{aligned} ^{0}\dot{\boldsymbol{\omega}}_i &= {^{0}\dot{\boldsymbol{\omega}}_{i-1}} + {^{0}\dot{\mathbf{R}}_i}\dot{\boldsymbol{\theta}}_i \mathbf{Z}_0 + {^{0}\mathbf{R}_i}\ddot{\boldsymbol{\theta}}_i \mathbf{Z}_0 \\ &= {^{0}\dot{\boldsymbol{\omega}}_{i-1}} + {^{0}\boldsymbol{\omega}_i} \times {^{0}\mathbf{R}_i}\dot{\boldsymbol{\theta}}_i \mathbf{Z}_0 + {^{0}\mathbf{R}_i}\ddot{\boldsymbol{\theta}}_i \mathbf{Z}_0 \end{aligned} \quad (10)$$

$$\begin{cases} ^{0}\ddot{\mathbf{P}}_i = {^{0}\ddot{\mathbf{P}}_{i-1}} + {^{0}\dot{\boldsymbol{\omega}}_{i-1}} \times {^{0}\mathbf{R}_{i-1}}\,{^{i-1}\mathbf{P}_i} + {^{0}\boldsymbol{\omega}_{i-1}} \times {^{0}\dot{\mathbf{R}}_{i-1}}\,{^{i-1}\mathbf{P}_i} \\ \quad = {^{0}\ddot{\mathbf{P}}_{i-1}} + {^{0}\dot{\boldsymbol{\omega}}_{i-1}} \times {^{0}\mathbf{R}_{i-1}}\,{^{i-1}\mathbf{P}_i} + {^{0}\boldsymbol{\omega}_{i-1}} \times \left( {^{0}\boldsymbol{\omega}_{i-1}} \times {^{0}\mathbf{R}_{i-1}}\,{^{i-1}\mathbf{P}_i} \right) \\ ^{0}\ddot{\mathbf{P}}_i^{c} = {^{0}\ddot{\mathbf{P}}_i} + {^{0}\dot{\boldsymbol{\omega}}_i} \times {^{0}\mathbf{R}_i}\,{^{i}\mathbf{P}_i^{c}} + {^{0}\boldsymbol{\omega}_i} \times {^{0}\dot{\mathbf{R}}_i}\,{^{i}\mathbf{P}_i^{c}} \\ \quad = {^{0}\ddot{\mathbf{P}}_i} + {^{0}\dot{\boldsymbol{\omega}}_i} \times {^{0}\mathbf{R}_i}\,{^{i}\mathbf{P}_i^{c}} + {^{0}\boldsymbol{\omega}_i} \times \left( {^{0}\boldsymbol{\omega}_i} \times {^{0}\mathbf{R}_i}\,{^{i}\mathbf{P}_i^{c}} \right) \end{cases} \quad (11)$$

### 3.2 动力学建模

由于 AM 系统动力学相对复杂，目前研究中多做简化处理，而实际 AM 系统为一种树形无根多刚体系统。

本文采用递归牛顿-欧拉方程建立 AM 系统动力学模型，不同于以往文献中的相对坐标系表示法，本文动力学参数均为惯性坐标系 $\{W\}$ 下表示，以便梳理刚体间内力（矩）关系。旋翼平台的动力方程如下：

$$\begin{cases} ^{0}\mathbf{I}_A\,{^{0}\dot{\boldsymbol{\omega}}_A} + {^{0}\boldsymbol{\omega}_A} \times {^{0}\mathbf{I}_A}\,{^{0}\boldsymbol{\omega}_A} = {^{0}\mathbf{U}_\tau} - {^{0}\boldsymbol{\tau}_1} - {^{0}\mathbf{r}_A^{\mathrm{out}}} \times {^{0}\mathbf{f}_1} \\ m_A\,{^{0}\ddot{\mathbf{P}}_A} = {^{0}\mathbf{U}_f} - {^{0}\mathbf{f}_1} + \mathbf{G}_A \end{cases} \quad (12)$$

其中 $^{0}\mathbf{U}_f$、$^{0}\mathbf{U}_\tau$ 分别为旋翼产生的驱动力（矩），$^{0}\mathbf{f}_1$ 与 $^{0}\boldsymbol{\tau}_1$ 分别为旋翼平台对连杆 1 的力（矩），$^{0}\mathbf{r}_A^{\mathrm{out}}$ 为作用在连杆 1 的力臂向量，$m_A$ 与 $^{0}\mathbf{I}_A$ 分别为旋翼平台质量和相对 $\{W\}$ 的惯性张量，$\mathbf{G}_A$ 为旋翼机械臂重力向量。

结合(10)、(11)式，连杆 $i$ 的动力学方程为：

$$\begin{cases} {}^0\mathbf{I}_i {}^0\dot{\boldsymbol{\omega}}_i + {}^0\boldsymbol{\omega}_i \times {}^0\mathbf{I}_i {}^0\boldsymbol{\omega}_i = {}^0\boldsymbol{\tau}_i + {}^0\mathbf{r}_i^{\text{in}} \times {}^0\mathbf{f}_i - {}^0\boldsymbol{\tau}_{i+1} - {}^0\mathbf{r}_i^{\text{out}} \times {}^0\mathbf{f}_{i+1} \\ m_i^m {}^0\ddot{\mathbf{P}}_i^c = {}^0\mathbf{f}_i - {}^0\mathbf{f}_{i+1} + \mathbf{G}_i^m \\ {}^0\dot{\boldsymbol{\omega}}_i = {}^0\dot{\boldsymbol{\omega}}_{i-1} + {}^0\boldsymbol{\omega}_i \times {}^0\mathbf{R}_i \dot{\boldsymbol{\theta}}_i \mathbf{Z}_0 + {}^0\mathbf{R}_i \ddot{\boldsymbol{\theta}}_i \mathbf{Z}_0 \\ {}^0\ddot{\mathbf{P}}_i = {}^0\ddot{\mathbf{P}}_{i-1} + {}^0\dot{\boldsymbol{\omega}}_{i-1} \times {}^0\mathbf{R}_{i-1} {}^{i-1}\mathbf{P}_i + {}^0\boldsymbol{\omega}_{i-1} \times \left( {}^0\boldsymbol{\omega}_{i-1} \times {}^0\mathbf{R}_{i-1} {}^{i-1}\mathbf{P}_i \right) \\ {}^0\ddot{\mathbf{P}}_i^c = {}^0\ddot{\mathbf{P}}_i + {}^0\dot{\boldsymbol{\omega}}_i \times {}^0\mathbf{R}_i {}^i\mathbf{P}_i^c + {}^0\boldsymbol{\omega}_i \times \left( {}^0\boldsymbol{\omega}_i \times {}^0\mathbf{R}_i {}^i\mathbf{P}_i^c \right) \\ {}^0\mathbf{r}_i^{\text{in}} = {}^0\mathbf{P}_i - {}^0\mathbf{P}_i^c \\ {}^0\mathbf{r}_i^{\text{out}} = {}^0\mathbf{P}_{i+1} - {}^0\mathbf{P}_i^c \end{cases} \quad (13)$$

其中 ${}^0\mathbf{f}_i$、${}^0\boldsymbol{\tau}_i$ 分别为连杆 $i-1$ 对连杆 $i$ 的力（矩）（文本中 ${}^0\mathbf{f}_6$、${}^0\boldsymbol{\tau}_6$ 为末端连杆对外部环境作用力、矩），${}^0\mathbf{r}_i^{\text{in}}$、${}^0\mathbf{r}_i^{\text{out}}$ 分别为连杆 $i-1$、$i$ 对连杆 $i$、$i+1$ 的力臂向量，$m_i$ 与 ${}^0\mathbf{I}_i$ 分别为连杆质量和相对 $\{W\}$ 的惯性张量，$\mathbf{G}_i^m$ 为连杆重力向量。

由本文 2 节知，机械臂关节角速度 $\dot{\boldsymbol{\theta}}$、角加速度 $\ddot{\boldsymbol{\theta}}$ 为已知输入，并由(12)、(13)式得出，AM 系统动力学可建模为以 $5m+2$ 个 3 维向量为变量的二阶代数微分方程组（Differential - Algebraic Equations，DAEs），且 AM 动力学模型存在强非线性，旋翼平台与机械臂间存在动力学约束。**不同于引言文献中的简化建模，(6)-(13)式给出了 AM 子系统间精确动力学描述。**

## 4 全驱动态补偿 PID 控制

### 4.1 全向驱动 PID 控制

分析旋翼平台位姿控制问题，(12)式 ${}^0\mathbf{I}_A$ 项由旋翼平台姿态矩阵决定，

$$ {}^0\mathbf{I}_A = {}^0\mathbf{R}_A {}^A\mathbf{I}_A {}^0\mathbf{R}_A^T \quad (14)$$

${}^A\mathbf{I}_A$ 为旋翼平台相对 $\{A\}$ 系的惯性张量常数矩阵，因此旋翼姿态动力学存在时变性与非线性，其姿态控制相对复杂。为此本文变换(2)式为指数坐标形式，并依此建立姿态控制误差。定义斜对称矩阵：

$$\mathbf{s}(\mathbf{v}) = \begin{bmatrix} 0 & -v_z & v_y \\ v_z & 0 & -v_x \\ -v_y & v_x & 0 \end{bmatrix}, \mathbf{v} = [v_x, v_y, v_z]^T \quad (15)$$

${}^0\mathbf{R}_A$ 可由指数参数 $\langle \mathbf{v}_\omega, \theta_\omega \rangle$ 表示为

$${}^0\mathbf{R}_A = \mathbf{I}_{3\times3} + \mathbf{s}(\mathbf{v}_\omega) s\theta_\omega + \mathbf{s}(\mathbf{v}_\omega)^2 (1 - c\theta_\omega) \quad (16)$$

${}^0\mathbf{R}_A$ 可解释为旋翼平台绕以相对 $\{W\}$ 系的 $\mathbf{v}_\omega$ 向量为转轴方向旋转 $\theta_\omega$ 角度得到姿态矩阵。设期望姿态矩阵为 ${}^0\mathbf{R}_A^d$，当前姿态矩阵为 ${}^0\mathbf{R}_A^c$，两者间变换矩阵为：

$${}^0\mathbf{R}_A^e = {}^0\mathbf{R}_A^d \left({}^0\mathbf{R}_A^c\right)^T \quad (17)$$

且由 ${}^0\mathbf{R}_A^d$ 得到指数参数：

$$\begin{cases} \theta_\omega^e = \cos^{-1}\left( \frac{tr\left({}^0\mathbf{R}_A^e\right) - 1}{2} \right) \\ \mathbf{v}_\omega^e = \frac{1}{2\sin\theta_\omega^e} \begin{pmatrix} r_{32} - r_{23} \\ r_{13} - r_{31} \\ r_{21} - r_{12} \end{pmatrix} \end{cases} \quad (18)$$

进而可直观解释 ${}^0\mathbf{R}_A^c$ 到 ${}^0\mathbf{R}_A^d$ 是旋翼平台绕以相对 $\{W\}$ 系的 $\mathbf{v}_\omega^e$ 向量为转轴方向旋转 $\theta_\omega^e$ 角度得到 ${}^0\mathbf{R}_A^d$。指数参数将三次欧拉角变换的姿态矩阵等价为一次固定轴旋转变换，因此可通过转矩输入控制角加速度以直观控制变换矩阵。

本文先以 PID 控制作为旋翼平台全向控制方法，定义位姿误差为 $\mathbf{e} = \left[\mathbf{e}_f^T, \mathbf{e}_\tau^T\right]^T$，其中 $\mathbf{e}_\tau = \theta_\omega^e \cdot \mathbf{v}_\omega^e$，$\mathbf{e}_f = {}^0\mathbf{P}_A^d - {}^0\mathbf{P}_A^c$，${}^0\mathbf{P}_A^c$ 与 ${}^0\mathbf{P}_A^d$ 分别为旋翼平台相对 $\{W\}$ 系当前位置与期望位置，则旋翼平台全向位姿驱动 PID 控制的矩阵形式为：

$$\mathbf{u}_{\text{pid}}(t) = \mathbf{k}_p \mathbf{e}(t) + \mathbf{k}_i \int_0^t \mathbf{e}(\tau) d\tau + \mathbf{k}_d \dot{\mathbf{e}}(t) \quad (19)$$

考虑系统可视为一类二阶非线性不确定系统，因此可以参考[29]分析并调节参数。

### 4.2 动态内力(矩)补偿

由(12)-(13)看出系统动力学存在内力约束耦合，仅依靠 PID 控制不能满足控制要求。为此本文估计动力学方程中内力约束项并依此建立动态补偿量以降低内力约束影响。旋翼平台位姿控制的动态补偿量为：

$$\mathbf{u}_{\text{dc}} = \begin{bmatrix} {}^0\boldsymbol{\omega}_A \times {}^0\mathbf{I}_A {}^0\boldsymbol{\omega}_A + {}^0\boldsymbol{\tau}_1 + {}^0\mathbf{r}_A^{\text{out}} \times {}^0\mathbf{f}_1 \\ \mathbf{G}_A - {}^0\mathbf{f}_1 \end{bmatrix} \quad (20)$$

从上式看出 $\mathbf{u}_{\text{dc}}$ 由机械臂作用力决定，但机械臂作用力又与旋翼平台构成了复杂代数环，该问题也是综述文献简化处理的原因之一。

本文视动力学微分方程组为代数方程组，并求解 AM 内部约束力(矩)，进而得到动力学补偿。选

择 $\mathbf{X}_s = \left[ {}^0\dot{\boldsymbol{\omega}}_A^T, {}^0\ddot{\mathbf{P}}_A^T, {}^0\mathbf{f}^T, {}^0\boldsymbol{\tau}^T, {}^0\dot{\boldsymbol{\omega}}^T, {}^0\ddot{\mathbf{P}}^T, \left({}^0\ddot{\mathbf{P}}^c\right)^T \right]^T$ 为待解变量，则可将(12)、(13)联立并改写为矩阵形式($n=5$)：

$$\begin{bmatrix} \mathbf{M}_A \\ \mathbf{M}_M^1 \\ \cdots \\ \mathbf{M}_M^n \end{bmatrix} \mathbf{X}_s = \begin{bmatrix} \mathbf{b}_A \\ \mathbf{b}_M^1 \\ \cdots \\ \mathbf{b}_M^n \end{bmatrix} \quad (21)$$

方程求解过程中除 $\mathbf{X}_s$ 外均视为常数，则根据(12)、(13)整理出具体矩阵形式(其中 $e_*(\cdot)$ 为矩阵在方程位置的扩增矩阵)：

$$\begin{cases} \mathbf{M}_A = \begin{bmatrix} {}^0\mathbf{I}_A & \mathbf{0}_{3\times 3} & \mathbf{e}_1\left({}^0\mathbf{r}_A^{out} \times\right) & \mathbf{e}_1(\mathbf{I}_{3\times 3}) & \mathbf{0}_{3\times 3n} & \mathbf{0}_{3\times 3n} & \mathbf{0}_{3\times 3n} \\ \mathbf{0}_{3\times 3} & m_A \mathbf{I}_{3\times 3} & \mathbf{e}_1(\mathbf{I}_{3\times 3}) & \mathbf{0}_{3\times 3n} & \mathbf{0}_{3\times 3n} & \mathbf{0}_{3\times 3n} & \mathbf{0}_{3\times 3n} \end{bmatrix} \\ \mathbf{b}_A = \begin{bmatrix} \mathbf{U}_\tau - \boldsymbol{\omega}\mathbf{I}\boldsymbol{\omega}_A \\ \mathbf{U}_f + \mathbf{G}_A \end{bmatrix} \end{cases} \quad (22)$$

$$\begin{cases} \mathbf{M}_1 = \begin{bmatrix} \mathbf{0}_{3\times 3} & \mathbf{0}_{3\times 3} & \begin{matrix} \mathbf{e}_2\left({}^0\mathbf{r}_1^{out}\times\right) \\ -\mathbf{e}_1\left({}^0\mathbf{r}_1^{in}\times\right) \end{matrix} & \begin{matrix} \mathbf{e}_2(\mathbf{I}_{3\times 3}) \\ -\mathbf{e}_1(\mathbf{I}_{3\times 3}) \end{matrix} & \mathbf{e}_1\left({}^0\mathbf{I}_1\right) & \mathbf{0}_{3\times 3n} & \mathbf{0}_{3\times 3n} \\ \mathbf{0}_{3\times 3} & \mathbf{0}_{3\times 3} & \begin{matrix} \mathbf{e}_1(-\mathbf{I}_{3\times 3}) \\ +\mathbf{e}_2(\mathbf{I}_{3\times 3}) \end{matrix} & \mathbf{0}_{3\times 3n} & \mathbf{0}_{3\times 3n} & \mathbf{0}_{3\times 3n} & \mathbf{e}_1\left(m_1^{m0}\mathbf{I}_{3\times 3}\right) \\ -\mathbf{I}_{3\times 3} & \mathbf{0}_{3\times 3} & \mathbf{0}_{3\times 3n} & \mathbf{0}_{3\times 3n} & \mathbf{e}_1(\mathbf{I}_{3\times 3}) & \mathbf{0}_{3\times 3n} & \mathbf{0}_{3\times 3n} \\ \mathbf{0}_{3\times 3} & \mathbf{0}_{3\times 3} & \mathbf{0}_{3\times 3n} & \mathbf{0}_{3\times 3n} & \mathbf{e}_1(-\mathbf{s}_1^1) & -\mathbf{e}_1(\mathbf{I}_{3\times 3}) & \mathbf{e}_1(\mathbf{I}_{3\times 3}) \\ -\mathbf{s}_1^2 & -\mathbf{I}_{3\times 3} & \mathbf{0}_{3\times 3n} & \mathbf{0}_{3\times 3n} & \mathbf{0}_{3\times 3n} & \mathbf{e}_1(\mathbf{I}_{3\times 3}) & \mathbf{0}_{3\times 3n} \end{bmatrix} \\ \mathbf{b}_i = \begin{bmatrix} -\boldsymbol{\omega}\mathbf{I}\boldsymbol{\omega}_i \\ \mathbf{G}_i^m \\ \mathbf{h}_i \\ \mathbf{g}_i^2 \\ \mathbf{g}_i^2 \end{bmatrix} \end{cases} \quad (23)$$

$$\begin{cases} \mathbf{M}_i = \begin{bmatrix} \mathbf{0}_{3\times 3} & \mathbf{0}_{3\times 3} & \begin{matrix} \mathbf{e}_i\left({}^0\mathbf{r}_i^{out}\times\right) \\ -\mathbf{e}_i\left({}^0\mathbf{r}_i^{in}\times\right) \end{matrix} & \begin{matrix} \mathbf{e}_{i+1}(\mathbf{I}_{3\times 3}) \\ -\mathbf{e}_i(\mathbf{I}_{3\times 3}) \end{matrix} & \mathbf{e}_i\left({}^0\mathbf{I}_i\right) & \mathbf{0}_{3\times 3n} & \mathbf{0}_{3\times 3n} \\ \mathbf{0}_{3\times 3} & \mathbf{0}_{3\times 3} & \begin{matrix} \mathbf{e}_i(-\mathbf{I}_{3\times 3}) \\ +\mathbf{e}_{i+1}(\mathbf{I}_{3\times 3}) \end{matrix} & \mathbf{0}_{3\times 3n} & \mathbf{0}_{3\times 3n} & \mathbf{0}_{3\times 3n} & \mathbf{e}_i\left(m_i^{m0}\mathbf{I}_{3\times 3n}\right) \\ \mathbf{0}_{3\times 3} & \mathbf{0}_{3\times 3} & \mathbf{0}_{3\times 3n} & \mathbf{0}_{3\times 3n} & \begin{matrix} \mathbf{e}_i(\mathbf{I}_{3\times 3}) \\ -\mathbf{e}_{i-1}(\mathbf{I}_{3\times 3}) \end{matrix} & \mathbf{0}_{3\times 3n} & \mathbf{0}_{3\times 3n} \\ \mathbf{0}_{3\times 3} & \mathbf{0}_{3\times 3} & \mathbf{0}_{3\times 3n} & \mathbf{0}_{3\times 3n} & \mathbf{e}_i(-\mathbf{s}_i^1) & -\mathbf{e}_i(\mathbf{I}_{3\times 3}) & \mathbf{e}_i(\mathbf{I}_{3\times 3}) \\ \mathbf{0}_{3\times 3} & \mathbf{0}_{3\times 3} & \mathbf{0}_{3\times 3n} & \mathbf{0}_{3\times 3n} & \mathbf{e}_{i-1}(-\mathbf{s}_i^2) & \begin{matrix} \mathbf{e}_i(\mathbf{I}_{3\times 3}) \\ -\mathbf{e}_{i-1}(\mathbf{I}_{3\times 3}) \end{matrix} & \mathbf{0}_{3\times 3n} \end{bmatrix} \end{cases} \quad (24)$$

其中常数项为：

$$\begin{cases}{}^0\mathbf{I}_i^m = {}^0\mathbf{R}_i\,{}^i\mathbf{I}_i^{m0}\mathbf{R}_i^T \\ \boldsymbol{\omega}\mathbf{I}\boldsymbol{\omega}_i = {}^0\boldsymbol{\omega}_i \times {}^0\mathbf{I}_i^m\,{}^0\boldsymbol{\omega}_i \\ \mathbf{G}_i^m = m_i^m\mathbf{V}_g \\ \mathbf{h}_i = {}^0\boldsymbol{\omega}_i \times {}^0\mathbf{R}_i\dot{\boldsymbol{\theta}}_i\mathbf{Z}_0 + {}^0\mathbf{R}_i\ddot{\boldsymbol{\theta}}_i\mathbf{Z}_0 \\ \mathbf{s}_i^1 = \mathbf{s}\left(-{}^0\mathbf{R}_i\,{}^i\mathbf{P}_i^c\right) \\ \mathbf{s}_i^2 = \mathbf{s}\left(-{}^0\mathbf{R}_{i-1}\,{}^{i-1}\mathbf{P}_i\right) \\ \mathbf{g}_i^1 = g\left({}^0\boldsymbol{\omega}_i,{}^0\mathbf{R}_i,{}^i\mathbf{P}_i^c\right) \\ \mathbf{g}_i^2 = g\left({}^0\boldsymbol{\omega}_{i-1},{}^0\mathbf{R}_{i-1},{}^{i-1}\mathbf{P}_i\right) \\ g(\boldsymbol{\omega},\mathbf{R},\mathbf{P}) = \boldsymbol{\omega}\times(\boldsymbol{\omega}\times\mathbf{RP})\end{cases} \quad (25)$$

在控制迭代中计算矩阵参数并求解 $\mathbf{X}_s$，得到(20)所需补偿项估计。综合(12、19、20)式全向驱动旋翼平台动态补偿 PID(Dynamic Compensation PID, dc-PID)控制如下：

$$\mathbf{u} = \begin{bmatrix} m_A\mathbf{I}_{3\times 3} & \mathbf{0}_{3\times 3} \\ \mathbf{0}_{3\times 3} & {}^0\mathbf{I}_A \end{bmatrix}\mathbf{u}_{pid} + \mathbf{u}_{dc} \quad (26)$$

### 4.3 驱动力解耦

旋翼平台实际控制输出为旋翼桨转速，旋翼平台驱动力(矩)均由其产生，并且它们之间存在耦合关系。**为更好的分析与设计全向驱动平台，本文推导多旋翼驱动力矩与转速耦合矩阵的统一形式。**驱动力(矩)方向相对$\{A\}$不变，因此以下讨论均以$\{A\}$系表示。设 $n$ 个旋翼推进力相对$\{A\}$系的方向向量构成的矩阵为 $\mathbf{e}^F \in \Re^{3\times n}, \|\mathbf{e}_i^F\| = 1$，且推进力幅值 $f$ 近似正比于旋翼转速 $\Omega^2$ [30]：

$$f = k_f\Omega^2 \quad (27)$$

其中 $k_f$ 为推进系数，则旋翼 $i$ 相对$\{A\}$系推进力为：

$${}^A\mathbf{F}_i = \mathbf{e}_i^F k_f\Omega_i^2 \quad (28)$$

其中 $k_{tf}$ 为推进力系数，相对$\{A\}$系合力为：

$${}^A\mathbf{F} = \sum_{i=1}^N \mathbf{e}_i^F k_f\Omega_i^2 \quad (29)$$
$$= k_f\mathbf{e}^F\boldsymbol{\Omega}$$

其中 $\boldsymbol{\Omega} \triangleq \left[\Omega_1^2, \Omega_2^2, \ldots, \Omega_n^2\right]^T$。

旋翼旋转产生反转矩与其产生的推进力同轴，且幅值亦近似正比于 $\Omega^2$，设旋翼 $i$ 相对$\{A\}$系力作用点向量为 ${}^A\mathbf{r}_i \in \Re^3$，则力矩 ${}^A\boldsymbol{\tau}_i$ 表示为：

$$\boldsymbol{\tau}_i = \mathbf{r}_i \times \mathbf{e}_i^F k_f\Omega_i^2 + d_i k_\tau \mathbf{e}_i^F\Omega_i^2 \quad (30)$$

其中 $d_i \in \{-1,1\}$ 为方向系数，$k_\tau$ 为转矩系数，整理得：

$$\begin{aligned}{}^A\boldsymbol{\tau}_i &= {}^A\mathbf{r}_i \times \mathbf{e}_i^F k_f\Omega_i^2 + d_i k_\tau \mathbf{e}_i^F\Omega_i^2 \\ &= \mathbf{s}\left(k_f\,{}^A\mathbf{r}_i\right)\mathbf{e}_i^F\Omega_i^2 + d_i k_\tau\mathbf{I}_{3\times 3}\mathbf{e}_i^F\Omega_i^2 \quad (31) \\ &\triangleq \mathbf{s}_\tau\left({}^A\mathbf{r}_i, d_i, k_f, k_\tau\right)\mathbf{e}_i^F\Omega_i^2\end{aligned}$$

其中，

$$\mathbf{s}_\tau(\mathbf{r}, d, k_f, k_\tau) = \begin{bmatrix} k_\tau d & -k_f r_3 & k_f r_2 \\ k_f r_3 & k_\tau d & -k_f r_1 \\ -k_f r_2 & k_f r_1 & k_\tau d \end{bmatrix} \quad (32)$$

则$\{A\}$系下旋翼产生的力矩为：

$$\boldsymbol{\tau} = \mathbf{M}_\tau\boldsymbol{\Omega} \quad (33)$$

$$\mathbf{M}_\tau = \left[\mathbf{M}_\tau^1, \mathbf{M}_\tau^2, \ldots, \mathbf{M}_\tau^n\right], \mathbf{M}_\tau^i = \mathbf{s}_\tau(\mathbf{r}_i, d_i, k_{tf}, k_{ct})\mathbf{e}_i^F$$

综合(29)与(33)式旋翼旋转产生的广义力为：

$$\begin{bmatrix} {}^A\mathbf{F} \\ {}^A\boldsymbol{\tau} \end{bmatrix} = \begin{bmatrix} k_f\mathbf{e}^F \\ \mathbf{M}_\tau \end{bmatrix}\boldsymbol{\Omega} \quad (34)$$

**(34)式为多旋翼驱动耦合矩阵的一般形式，[31, 32]中的耦合矩阵形式均可视为上式的一般形式，因此本文建立的耦合矩阵一般形式更便于分析与设计旋翼平台。**

相对$\{W\}$系控制输出与旋翼转速关系为：

$$\begin{bmatrix} {}^0\mathbf{U}_f \\ {}^0\mathbf{U}_\tau \end{bmatrix} = {}^0\mathbf{R}_A\begin{bmatrix} k_f\mathbf{e}^F \\ \mathbf{M}_\tau \end{bmatrix}\boldsymbol{\Omega} \quad (35)$$

电机输出功率限制了转速范围，$0 \le \Omega \le \Omega_{max}$，其中 $\Omega_{max}$ 为最大转速，因此控制力(矩)对应旋翼转速的求取为带约束的线性最优化问题，本文采用内点法处理(Matlab 仿真中采用 lsqlin 并结合 lsqnonneg 函数)。

## 5 仿真与分析（Simulations and Analysises）

### 5.1 仿真系统介绍

采用 Matlab2016b 搭建仿真系统，如图 2 所示。考虑仿真时耗，下文仅展示简化 3D 模型，且模型仅反映位姿、尺寸比例关系，与内在动力学性能无关。仿真采用一阶惯性滤波与限幅操作以提升噪声抑制能力并使控制输出在功率限制范围内，作者开源本文仿真代码。

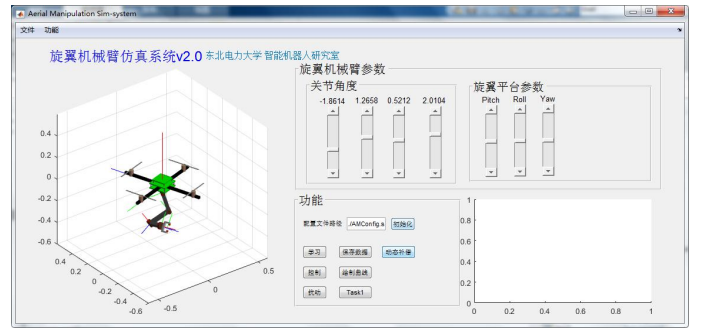

图 2 AM Matlab 仿真系统界面
Fig.2 Interface of AM simulation base on Matlab

## 5.2 位置控制测试

选取 4 种机械臂位型以全面测试分析本文方法，如图 3 所示。4 种条件的 $\{A\}$ 系质心向量 ${}^0\mathbf{P}_A$ 初始值与期望值均分别为 $[9,9,9]^T$ 与 $[10,10,10]^T$ （单位 m），期望 Euler 角均为 $[0,0,0]^T$（单位°）。

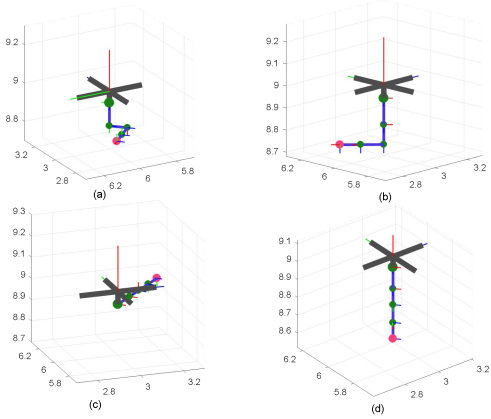

图 3 全向驱动控制仿真机械臂位型示意图
Fig.3 illustration of 4 types of configurations of manipulator for full-actuated control simulations

为不失一般性，测试中施加不同值于初始仿真参数，表 2 位置测试仿真参数描述。

表 2 仿真参数描述
Tab.2 Simulation Paramters Description

| 编号 | 初始欧拉角/° | 初始关节角度/° | 噪声与扰动 | 参数不确定范围 |
|---|---|---|---|---|
| 1 | -10,-10,-10 | -90,-90,0,0 | ±25mm, ±3° | ±10% |
| 2 | 15,-20,10 | -90,0 0 0 | ±25mm, ±3° | ±10% |
| 3 | 20,10,-15 | 0,0,-90,0 | ±25mm, ±3° | ±10% |
| 4 | -10,20,15 | -90,0,-90,0 | ±25mm, ±3° | ±10% |

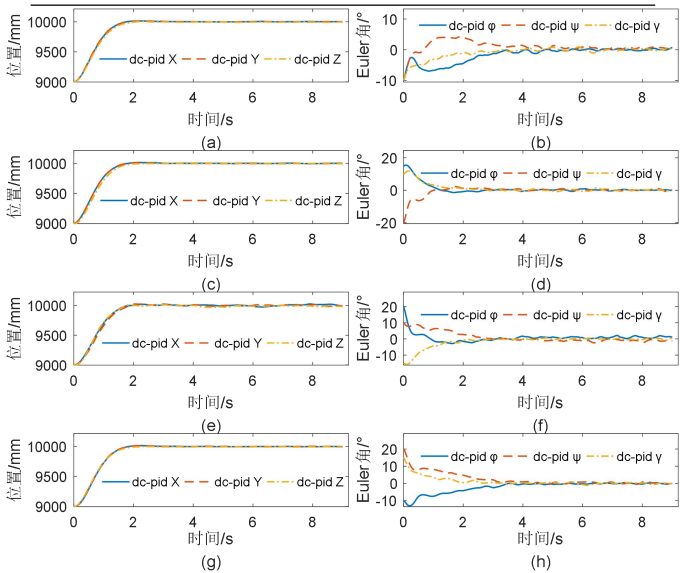

图 4 不同初始参数的 dc-PID 全向驱动位置控制结果
Fig.4 Results of fully-actuated control to posistions in various types of initial parameters using dc-PID

图 4 全向驱动位置控制结果，从中得出动态补偿 PID 控制能使系统以微超调(<0.5%)快速(调节时间<2s)收敛于期望位置，对于不同初始姿态系统均可收敛至期望姿态，且调节过程为 6 自由度同步执行，即为全向驱动控制。**因动态补偿作用使具有强非线性与耦合性的 AM 系统无明显震荡，同时一阶惯性滤波器作用使控制曲线平滑，对扰动与噪声具有较好的抑制能力。**受机械臂位型与仿真系统动力极限制约，位型 3 的控制曲线 ((e),(f))抖动较大，但仍在满意范围。

利用上述 4 位型仿真测试倾斜悬停控制，即在位置不变条件下使旋翼平台姿态具有倾角并稳定悬停。

4 位型初始与期望值均为 $[9,9,9]^T$ (单位 m)，初始姿态 Euler 角均为 $[0,0,0]^T$，期望值分别为 $[10,10,10]^T$、$[20,20,20]^T$、$[15,15,0]^T$、$[15,15,0]^T$ （单位°）。扰动与噪声、参数不确定范围同表 2。

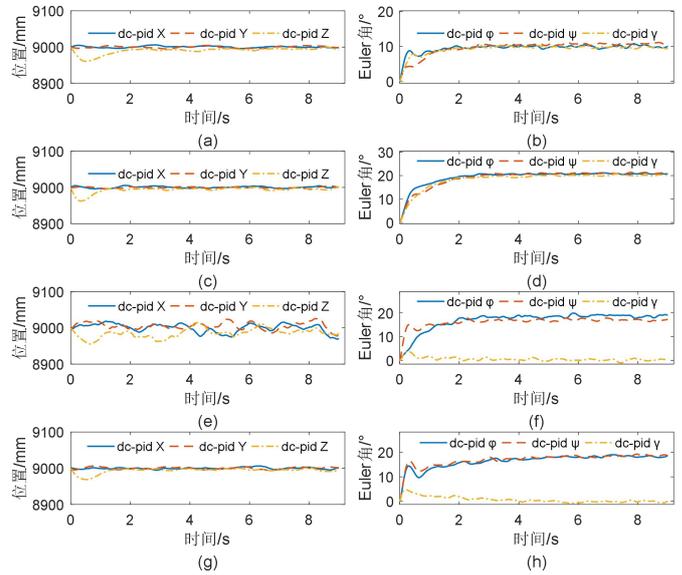

图 5 不同初始参数的 dc-PID 全向驱动倾斜悬停控制结果
Fig. 5 Results of fully-actuated control to hover with angles of inclination in various types of initial parameters using dc-PID

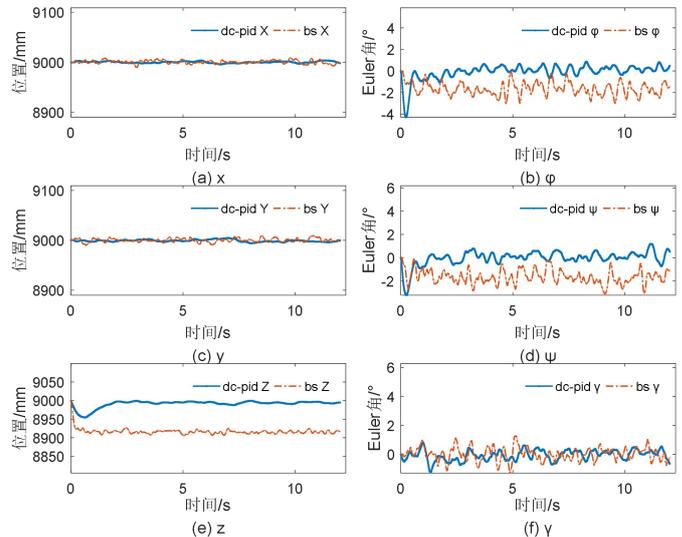

图 6 dc-PID 与 Backstepping 关节运动下悬停控制结果比较
Fig. 6 Simulation result comparisions of hover control between using dc-PID and Backstepping with rotation of arm-joint

图 5 为上述 4 位型对应表 2 初始条件的倾斜悬停控制结果。从中得出，4 位型均能在 2s 内收敛与期望倾角，X、Y 方向一直处于稳定，在初始阶段由于姿态调节的非线性与功率限制因素，Z 轴位置均出现＜25mm 的抖动，但扔能在 2s 内调整为期望位置。位型 3 的曲线抖动扔较大，结果与原因与上一测试一致。

本测试更重要意义在于，验证了本文设计的全向驱动 AM 系统能够完成传统系统理论上无法完成的带有倾斜角度的稳定悬停控制。

### 5.3 空中作业任务仿真

实际空中作业伴随机械臂运动，本节模拟两种空中作业任务，仿真测试本文设计系统。与[17, 18]等采用的 Backstepping 算法比较，分析提出的 dc-PID 控制性能。

任务 1 令 AM 系统旋翼平台初始位姿为 $[9m, 9m, 9m, 0°, 0°, 0°]$ 并稳定于此，机械臂关节角度为 $[-90° \ 0° \ -45° \ 0°]$，关节 3 以 $\dot{\theta}_3 = 0.5\sin(t)$ 速度运动，其他条件参数同上节。

图 6 为仿真结果，从中得出在机械臂运动、扰动噪声与参数不确定性条件下，dc-PID 与 Backstepping 均在受初始条件冲击下达到稳定状态。**但 Backstepping 结果抖动范围较大，对噪声的抑制能力弱于 dc-PID，且在 Z、$\varphi$、$\psi$ 较 dc-PID 存在较大稳定误差。**

任务 2 在上任务基础上增加了旋翼平台运动，令从任务 1 初始姿态调至 $[9.3m, 9.3m, 9.3m, 20°, 20°, 0°]$，其他条件参数同上。

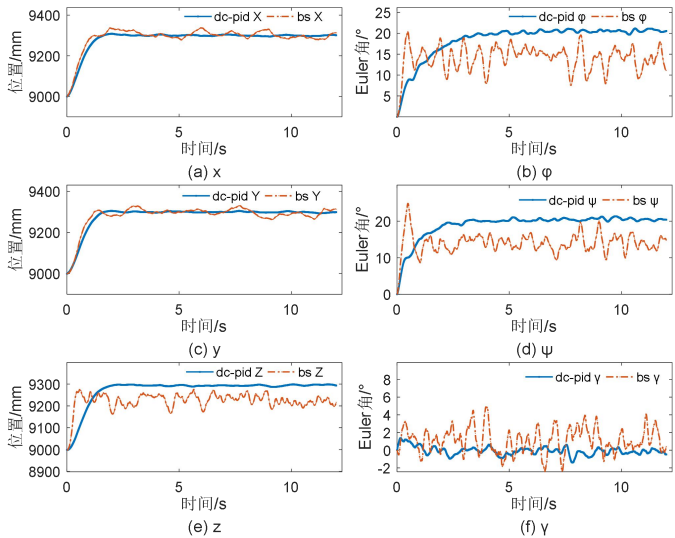

图 7 dc-PID 与 Backstepping 关节运动下全向驱动控制结果比较
Fig. 7 Simulation result comparisions of fully-actuated control between using dc-PID and Backstepping with rotation of arm-joint

图 7 为任务 2 仿真比较结果，除有与任务 1 的一致结论外，又得出由于任务难度提升，**Backstepping 虽亦可达稳定，但被控量的抖动范围较任务 1 明显增大，而 dc-PID 抖动范围则较小且两任务无明显差异。**

两任务仿真表明，由于动态补偿与滤波等作用，**dc-PID 方法较 Backstepping 控制有更好的动态与稳定控制性能。**

本节仿真结果表明，本文设计的全向驱动 AM 系统能实现传统 AM 系统理论上无法实现的全向驱动控制，提出的 dc-PID 控制方法对全驱控制问题具有良好的控制性能。

## 6 实验（Experiments）

本文设计的全向旋翼机械臂样机如图 8 所示，图中 3D 打印设计驱动力与 {A} 坐标系夹角 $\beta_i$ 均为 45°采用基于 STM32F104 的自研控制器实现旋翼与机械臂控制，姿态检测采用 MPU6050 集成模块。

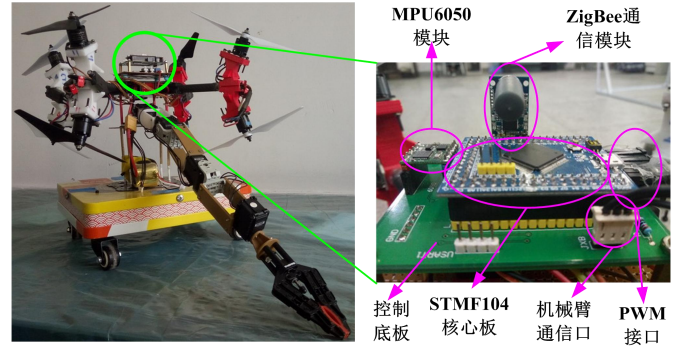

图 8 全向驱动 AM 系统原型机示意图
Fig. 8 illustration of the prototype of fully-actuated AM system

全向驱动旋翼机械臂控制实验前，以作者团队搭建的四旋翼平台实验测试 dc-PID 控制方法实用性与有效性，在 3D 打印制作机械臂上增加配重，模拟原型机机械臂负载。

图 9 为带有机械臂运动的 AM 系统 dc-PID 控制实验视频图列，从中看出在机械臂运动过程中四旋翼平台姿态能保持稳定，无较大改变，进而验证了 dc-PID 在实际控制中的实用性与有效性。

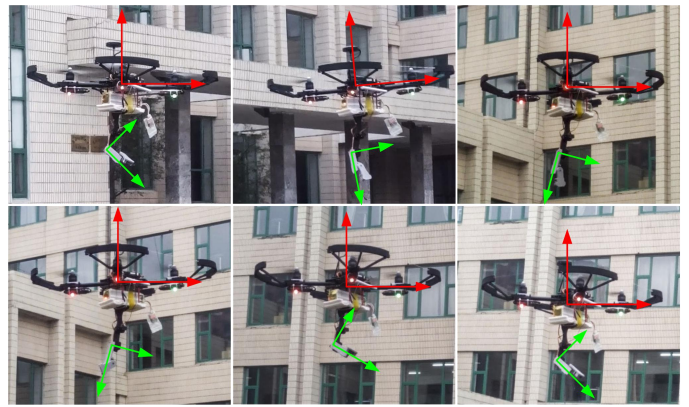

图 9 带有机械臂运动的 AM 系统 dc-PID 控制实验视频图列
Fig. 9 Images sequences from experimental vedio of AM system using dc-PID with motion of arm joint

## 7 结论与展望（Conclusion）

**通过以上论述与仿真、实验得出以下结论：**

1) 本文提出的全向驱动旋翼机械臂系统能够实现传统 AM 系统无法完成空中操作，特别是倾斜悬停、6 自由度同步控制等；

2) 本文建立了精确的 AM 系统内部动力学方程，明确揭示了 AM 内部动力学关系，并为后续控制提供了补偿依据；

3) 本文推导的耦合矩阵给出了多旋翼转速与驱动力(矩)的统一表达形式，提升了设计分析效率，并有助于驱动解算；

4）提出的 dc-PID 控制方法能够实现本文系统的全向驱动控制，并较 Backstepping 方法具有更好的控制精度与扰动抑制能力。

未来工作方向简述如下：
1) 优化样机，完成室内全驱控制测试；

2) 分析交互外力估计与环境交互方法；
3) 尝试其他控制算法以提升控制性能。

**作者简介：**

马　乐（1983--），男，博士，副教授。研究领域：旋翼机械臂系统与控制，机器人行为学习，机器人视觉。

XXXXXX（出生年--），性别，学位，职